\documentclass{article}
\usepackage{arxiv}
\usepackage[T1]{fontenc}
\usepackage[utf8]{inputenc} % allow utf-8 input
\usepackage{booktabs}       % professional-quality tables
\usepackage{amsfonts}       % blackboard math symbols
\usepackage{nicefrac}       % compact symbols for 1/2, etc.
\usepackage{microtype}      % microtypography
\usepackage{lipsum}
\usepackage{amsmath, amssymb}
\usepackage{multirow}
\usepackage{graphicx}
\usepackage{hyperref}
\usepackage{color}

\title{Local Interpretability of Random Forests for Multi-Target Regression}

\author{
  Avraam Bardos\\
    Aristotle University of Thessaloniki,\\ 54636, Greece\\
    \texttt{ampardosl@csd.auth.gr}\\
    \And
  Nikolaos Mylonas\\
    Aristotle University of Thessaloniki,\\ 54636, Greece\\
    \texttt{myloniko@csd.auth.gr}\\
    \And  
  Ioannis Mollas\\
    Aristotle University of Thessaloniki,\\ 54636, Greece\\
    \texttt{iamollas@csd.auth.gr}\\
    \And
  Grigorios Tsoumakas\\
    Aristotle University of Thessaloniki,\\ 54636, Greece\\
    \texttt{greg@csd.auth.gr}\\
}

\begin{document}
\maketitle      
\begin{abstract}

Multi-target regression is useful in a plethora of applications. Although random forest models perform well in these tasks, they are often difficult to interpret. Interpretability is crucial in machine learning, especially when it can directly impact human well-being. Although model-agnostic techniques exist for multi-target regression, specific techniques tailored to random forest models are not available. To address this issue, we propose a technique that provides rule-based interpretations for instances made by a random forest model for multi-target regression, influenced by a recent model-specific technique for random forest interpretability. The proposed technique was evaluated through extensive experiments and shown to offer competitive interpretations compared to state-of-the-art techniques.

\end{abstract}

\section{Introduction}
The problem of multi-target regression, has gained popularity in the last few years, and involves predicting two or more continuous values, similar to multi-label classification~\cite{waegeman2019multi}. Given an input dataset with features and corresponding target outputs, the goal is to estimate the target vector based on the input. Multi-target regression has several applications such as air quality monitoring~\cite{air_app} and electronic health record analysis~\cite{health_app}.

There are two main approaches to multi-target regression: problem transformation and algorithm adaptation. Problem transformation involves transforming a multi-target problem into multiple independent single-target problems, while algorithm adaptation adapts a specific method to handle multi-target regression data. Techniques falling into the latter category include predictive clustering trees (PCT)~\cite{pcts}, fitted rule ensembles (FIRE)~\cite{fire}, and random forests (RF)~\cite{rf}.

Interpretability provides reasoning behind the decisions of machine learning models. Most machine learning applications require interpretability, especially when they can affect human life directly (health applications) or indirectly (economic applications)~\cite{rc}. Interpretability in the context of multi-target regression is a scarcely explored topic, with very few techniques being developed for this task. The aforementioned PCTs are inherently interpretable, similar to traditional decision trees. Furthermore, FIRE is also inherently interpretable due to the small number and length of the rules it produces. On the other hand, RF is not inherently interpretable, and model-specific interpretability techniques for multi-target regression have not yet been developed.

Various techniques that are model-agnostic can be used to provide rule-based interpretations for single-label classification and regression tasks. Such techniques include RuleFit~\cite{rulefit}, Anchors~\cite{anchors}, and LORE~\cite{lore}. However, they cannot be used for multi-target regression. Additionally, a local model-agnostic technique called MARLENA~\cite{marlena} has been designed specifically for multi-label learning problems. 

This work addresses the challenge of interpretability of RF in the context of multi-target regression, by introducing a technique designed to explain predictions of specific instances, called XMTR (eXplainable Multi Target Regression random forests). Our technique is based on a recent technique for local model-specific interpretability of RF, called LionForests (LF), which can accommodate both single-label and multi-label tasks~\cite{lf,multilf}. XMTR provides rule-based interpretations for the predictions of the underlying model. To evaluate our method, we conducted quantitative, qualitative and scalability experiments, which demonstrate that XMTR offers competitive interpretations to those of state-of-the-art techniques. Overall, our results show that XMTR is a practical solution for enhancing the interpretability of RF in multi-target regression tasks.

\section{Explainable multi-target regression}

This work contributes a technique for addressing the interpretability of RF models for multi-target regression by proposing a modification to LF. First, we will present some fundamental concepts necessary for both LF and XMTR. Then, we will present the main three steps of LF, which include: a) path extraction, b) path reduction, and c) rule composition, and the modifications we are introducing.

LF provides a single rule, which explains the decision of a particular instance, without affecting the structure of the model. Through path reduction, it is possible to reduce the number of features in the rule, as well as to widen the feature ranges, making the interpretation more comprehensive to the end user. A rule provided by LF is formulated as a set of feature ranges. An example of such a rule is ``if $3 \leq f_2 \leq 8$ and $7 \leq f_3 \leq 12$ then $t_i$'', where a model predicted target $t_i$, based on features $f_2$ and $f_3$, among the available ones.

\paragraph{Path conclusiveness.}
\textit{Conclusiveness}, is a vital property of rules~\cite{lf}. A rule is considered to be conclusive if the prediction of the model being interpreted for the given instance remains the same, both when the values of the excluded features ($f_1$ and $f_4$ in our example) are modified arbitrarily and when the values of the included features ($f_2$ and $f_3$ in our example) are modified within the specified ranges.  

\paragraph{Allowed error.}
The concept of \emph{allowed error}~\cite{lf}, enables the formation of conclusive rules in tasks like regression. With \emph{allowed error}, it is possible to reduce the number of paths and form a smaller rule, while guaranteeing a predictive error within the allowed limit. The default value for \emph{allowed error} is equal to the mean absolute error of the model, according to an evaluation procedure (e.g. cross-validation), however, it can also be user-defined.

\paragraph{Local error.}
\emph{Allowed error} will be compared to a \emph{local error}, $\emph{local error} = mae(preds,$ $ r\_preds)$. The predictions made by all the trees are $preds$. The $r\_preds$ refers to both the prediction made by an individual tree when its path is included, or the highest/lowest possible prediction that the tree could make (highest/lowest value among all the tree's leaves) when its path is excluded, depending on which value is farthest from the actual prediction. If we have an RF with $|T|$ trees and exclude $|E|$ trees by choosing to keep only $|K|$, the final rule may not account for the decisions made by the excluded ones. If a change is made to a feature that is not included in the final rule, it can lead to a non-conclusive rule. Thus, the final prediction should be adjusted as follows: $p'(x)=\frac{1}{|T|}(\sum_{k\in K}^{}p_{k}(x)+\sum_{e\in E}^{}p_{e}(x))$, where $p_{k}(x)$ is the actual prediction of the tree $k$ and $p_{e}(x)$ is the minimum/maximum value the tree $e$ can predict.

\paragraph{Path extraction.}
The initial step is to perform a \textit{path extraction} procedure, which aims to collect all the paths taken by the trees in the RF that the given instance follows to obtain its predicted values. During this process, each tree is traversed, and the path that the instance takes from the root to the leaf node to arrive at its prediction is tracked. While following each path, the decisions of each internal node are examined, and the feature used for the split at that node is recorded, along with its threshold value. Hence, for each tree, a path is produced that includes all the features that were used, as well as their minimum and maximum threshold values, and the predicted value of the instance.

\paragraph{Path reduction.} 
Once the features and their ranges for each path are gathered, the subsequent crucial step is \textit{path reduction}. The purpose of this step is to exclude paths that are not essential to the decision. In~\cite{lf}, several reduction strategies are proposed, but here we focus solely on association rules reduction since it is suitable for regression tasks. Association rules are utilized to assign a confidence score to each path, and the features with lower confidence scores are gathered in an empty feature set. The iterative process of enriching the feature set continues until the number of paths, whose conjunctions of features are all included in the feature set, produces a \textit{local error} smaller than the \textit{allowed error}.

\paragraph{Rule composition.} 
The last step is to combine all knowledge acquired from the earlier steps, and create a single rule expressed in natural language. The process of \textit{rule composition} includes two steps: a) traverse all reduced paths, and b) for each feature appearing in the paths, keep track of their minimum and maximum values among the reduced paths (if included). This way, we create a conclusive rule by only including the necessary features to the prediction with respect to the predefined \emph{allowed error}. 

\begin{figure}[ht]
\centerline{\includegraphics[width=1\textwidth]{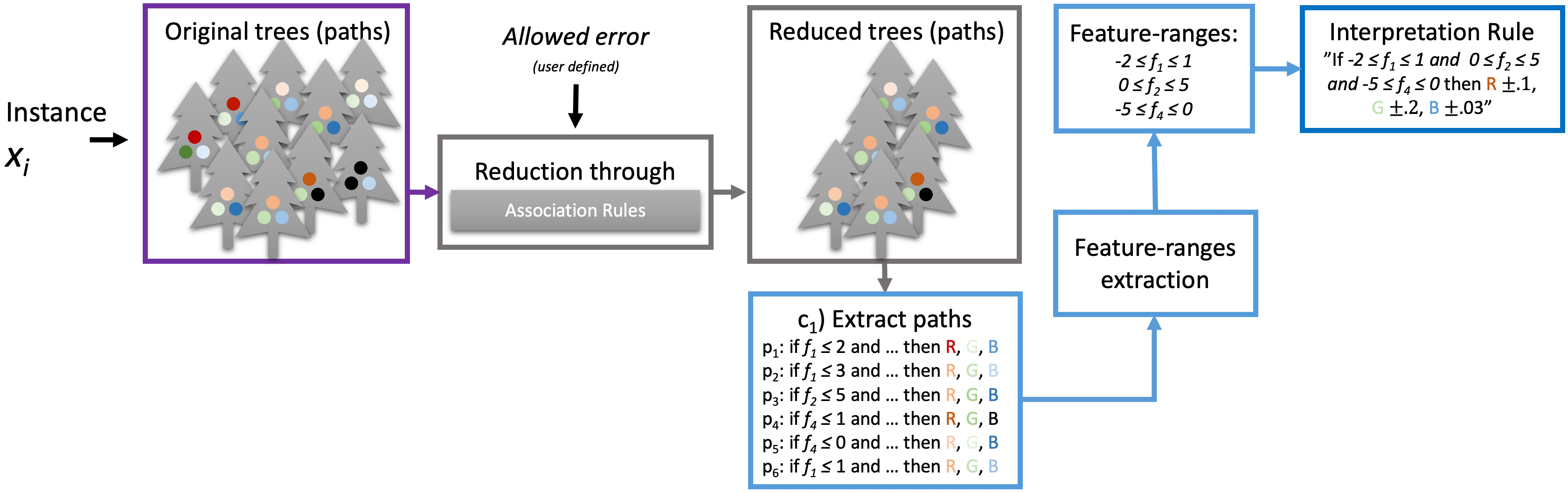}}
\caption{Workflow of the proposed approach}
\label{fig:workflow}
\end{figure}

\subsection{Adaptation for multi-target regression}
The goal of this work, through this adaptation, is to introduce a modification of LF to interpret multi-target regression RF models, named XMTR. XMTR provides a single rule explaining the whole target set. One such example rule, can be seen in the end of the workflow of Fig.~\ref{fig:workflow}. 

Among the available path reduction strategies of LF, we selected association rules for two reasons. The first one concerns its ability to reduce the paths with the goal of reducing the number of features, as well. This is crucial because the generated rules will only contain the necessary features. Second, it is the only strategy that can be modified to work for multi-target regression tasks. The goal is to produce a rule that explains the entire prediction, rather than a single specific target at a time.

Therefore, XMTR modifies the selected path reduction strategy to consider multiple allowed errors, one for each target. Then the local errors are compared based on two schemes. The first one compares their average to a globally shared allowed error, while the second one uses multiple allowed errors, each tailored to a specific target. The first scheme enables faster and simpler rule production, as the user defines only one error threshold. The second scheme provides greater flexibility, as the user can prioritize certain targets by setting a less strict allowed error for less important targets and a stricter one for desired targets.

\section{Experiments}
In this section, we first present the setting of our experiments, including a summary of the datasets we examined, the metrics we used and the competitors we compared. We then present the results of both the quantitative and qualitative experiments we conducted. Finally, we include a scalability analysis, based on the \textit{allowed error} parameter of XMTR. The datasets as well as the code for our technique and experiments are available in our XMTR repository \url{https://github.com/intelligence-csd-auth-gr/XMTR.git}.

\subsection{Experimental setup}

The datasets used in our experiments are Slump~\cite{mtrdata}, Andro~\cite{mtrdata}, and Facebook metrics~\cite{fm}, all of them concerning multi-target regression problems. These datasets have $\{103, 49, 500\}$ examples, $\{7, 30, 14\}$ features, and $\{3, 6, 4\}$ targets respectively. No pre-processing took place for all datasets besides Facebook metrics, where missing values were replaced with $0$, and categorical features were excluded, as in this work we do not address this issue.

We employed global (GS) and local (LS) decision tree surrogates, along with MARLENA (MA), in our study. As MARLENA was originally intended for multi-label interpretability, we had to make some adjustments to make it suitable for our multi-target regression task. In specific, we converted the surrogate model of the technique into a regressor, and modified the rules exported to include the prediction of the surrogate. These modifications were essential to achieve multi-target regression interpretability.

We used three metrics for our evaluation. The first one is {\em coverage}, which quantifies the amount of instances covered by a rule. Higher values indicate better performance for this metric. The second one is {\em rule precision}, which quantifies the number of correctly covered instances. Lower values indicate better performance, as we measure MAE. The last one is {\em rule length}, which quantifies the number of conjunctions present inside the rule. Depending on the user and case, lower or higher values may be preferred.

\subsection{Quantitative results}

The RF models' parameters were selected through a grid search process, and their values can be located in our repository at \href{https://github.com/intelligence-csd-auth-gr/XMTR.git}{XMTR}. However, we report the RF models' performance in MAE for each dataset; $1.7311$ for Slump, $1.3326$ for Andro, and $0.1008$ for Facebook metrics. Our quantitative experiments also involve three distinct \emph{allowed error} values for each dataset, which determine the maximum permissible amount of error for each rule concerning the target values. These \textit{allowed error} values are denoted with \#1, \#2, and \#3 and correspond to $0.2, 0.25,$ and $0.5$ for Slump, $0.55, 0.6,$ and $0.7$ for Andro, and, finally, $0.1, 0.3,$ and $0.5$ for Facebook metrics, respectively. While a plethora of \textit{allowed error} values was tested, this paper showcases only the ones that exhibit noticeable changes in the performance of MAE and rule length metrics, due to space limitations.

The data is partitioned using a 10-fold cross-validation strategy, and the training set is employed to initialize XMTR and its competitors. For every test instance, we generate interpretations and compute the metrics. Finally, we average these metrics across all instances, obtaining the final value for the corresponding metric.

Table~\ref{quantitative} presents the results. In terms of coverage, GS and LS had the higher performance, with XMTR and MA being lower. This means that the rules produced by the latter two techniques are more specific towards the instance. Concerning MAE, XMTR outperformed all other competitors for the \#1 and \#2 \textit{allowed errors}, but has worse performance for the \#3 which corresponds to higher \textit{allowed error} values in all three datasets. This is expected, as giving the technique a higher possible error value, would result in greater reduction, leading to more general rules, with bigger errors. The shortest rules were generated by LS with XMTR providing the lengthier ones. This is due to the rules provided by XMTR being more specific to the instance. Finally, higher \textit{allowed error} values resulted in shorter rules, which is a consequence of the greater path reduction discussed earlier.

\begin{table}[ht]
\centering
\begin{tabular}{rccccccccc}
\cline{2-10}
           & \multicolumn{3}{c}{Coverage} & \multicolumn{3}{c}{MAE} & \multicolumn{3}{c}{Rule Length} \\ \cline{2-10} 
Technique  & Slump    & Andro   & Facebook       & Slump    & Andro   & Facebook     & Slump    & Andro   & Facebook        \\ \hline
XMTR (\#1) & 0.09    & 0.13    & 0.04    & 0.09   & 0.06   & 0.00  & 6.8      & 21.4      & 7.9      \\
XMTR (\#2) & 0.09    & 0.17    & 0.04    & 0.17   & 0.25   & 0.02  & 6.7      & 20.4      & 7.6      \\
XMTR (\#3) & 0.10     & 0.20    & 0.05    & 1.46   & 0.68   & 0.28  & 4.9      & 18.1      & 5.6      \\
GS         & 0.20     & 0.35    & 0.19    & 0.22   & 0.34   & 0.02  & 3.9      & 4.3       & 3.6      \\
LS         & 0.23     & 0.34    & 0.19    & 0.30   & 0.40   & 0.03  & 2.9      & 3.5       & 2.7      \\
MA    & 0.11     & 0.23    & 0.09    & 0.16   & 0.21   & 0.02  & 4.8      & 6.9       & 4.7      \\ \hline
\end{tabular}
\caption{Results of the quantitative experiments comparing XMTR, GS, LS, and MA, regarding Slump, Andro, and Facebook metrics datasets, across the coverage, MAE, and rule length metrics}
\label{quantitative}
\end{table}

\subsection{Qualitative results}

We randomly selected an instance from the Slump dataset to conduct qualitative experiments, presenting the XMTR and MA rules. The dataset was selected due to the small number of features, allowing for better visualization. For the selected instance, the corresponding RF model predicted $7.9$ for Slump, $7.2$ for Flow, and $4.9$ for Compressive\_Strength targets.

\begin{quote}
    \textit{\textbf{XMTR:} if $207 \le Water \le 208$ \& $100 \le Slag \le 107$ \& $708 \le Coarse\_Aggr \le 753$ \& $137 \le Cemment \le 151$ \& $126 \le Fly\_ash \le 141$ then $Slump: 7.9^{\pm0.8}$, $Flow: 7.2^{\pm0.7}$, $Compressive\_Strength: 4.9^{\pm0.2}$}
\end{quote}

\begin{quote}
\textit{\textbf{MA:} if $185 < Water \le 209$ \& $Slag \le 144$ \& $Coarse\_Aggr \le 843$ \& $80 < Fly\_ash \le 198$ \& $ Fine\_Aggr \le 885$ then $Slump: 7.7$, $Flow: 6.8$, $Compressive\_Strength: 4.8$}
\end{quote}

Both rules included 5 of the 7 features. XMTR excluded $SP$ and $Fine\_Aggr$, while MA excluded $SP$ and $Cemment$. Changing the value of $SP$ had no impact on prediction, but modifying $Cemment$ from $145$ to $136$ (1 lower than the range suggested by XMTR) does affect it. In contrast, adjusting $Fine\_Aggr$ from $883$ to $640$ (the minimum value found in the train set) does not affect the prediction. Therefore, XMTR correctly excluded the two features that do not affect the prediction, while MA mistakenly excluded $Cemment$, which is a feature that appears to impact the target.

\subsection{Allowed error and scalability}

We aim to measure the effect of \textit{allowed error}, in the time response of XMTR. We anticipate a consistent increase in the time needed for rule generation in correlation to the value of \textit{allowed error}. Higher error values allows for greater path exclusion by XMTR, resulting in higher time response. We use two synthetic datasets created through \href{https://tinyurl.com/46vk4vax}{\textit{scikit-learn}}, one relatively small (D1) containing 500 instances, 10 features and 5 targets, as well as a larger one (D2) with 5000 instances, 50 features and 7 targets. Furthermore, the random forest model we used for this experiment consisted of 500 estimators, and we utilized five different \textit{allowed error} values. 

\begin{table}[ht]
\centering
\begin{tabular}{ccccc}
\cline{2-5}
              & \multicolumn{2}{c}{D1} & \multicolumn{2}{c}{D2} \\ \cline{2-5} 
\textit{allowed error} & Time          & \#Paths          & Time           & \#Paths          \\ \hline
0.05                   & 0.93          & 482               & 0.58           & 495               \\
0.1                    & 1.89          & 466               & 1.03           & 488               \\
0.15                   & 2.73          & 449               & 1.30           & 481               \\
0.2                    & 3.61          & 436               & 1.75           & 473               \\
0.25                   & 4.38          & 418               & 2.01           & 466               \\
0.3                    & 5.22          & 401               & 2.57           & 457               \\ \hline
\end{tabular}
\caption{Scalability analysis of XMTR}
\label{tab:scalability}
\end{table}

Table~\ref{tab:scalability}, presents the time required by XMTR to produce the rules, as well as the number of paths used for that rule out of 500 initial ones. It is visible through these results that our hypothesis holds true and higher error values allow for greater reduction and thus higher time responses. It is also important to mention that the total number of instances (500, 5000) did not have any significant influence on the time performance. Therefore, the choice of \textit{allowed error}, is the most critical to the time response of our method.

\section{Conclusions}
Our work tackled the challenge of interpretability in multi-target regression tasks for RF, which is a scarcely explored topic with important applications. We introduced a technique called XMTR, which modifies the existing LF approach for local model-specific interpretability of RF. Through quantitative and qualitative experiments, we demonstrated that XMTR outperforms current state-of-the-art techniques. Additionally, through a scalability analysis, we presented how XMTR's \textit{allowed error} parameter affects time response. Our research highlights the potential of XMTR in multi-target regression tasks and contributes to the advancement of interpretability research in this scarcely explored area.

\section*{Acknowledgments}
The research work was supported by the Hellenic Foundation for Research and Innovation (H.F.R.I.) under the ``First Call for H.F.R.I. Research Projects to support Faculty members and Researchers and the procurement of high-cost research equipment grant'' (Project Number: 514)

%\bibliographystyle{unsrt} % We choose the "plain" reference style
%\bibliography{bib} % Entries are in the refs.bib file

\end{document}